\documentclass[letterpaper, 10 pt, conference]{ieeeconf}

\IEEEoverridecommandlockouts

\IEEEoverridecommandlockouts  
\usepackage{amsmath}
\usepackage{siunitx}
\usepackage[pdftex]{graphicx}
\usepackage{times}
\usepackage{mathptmx}
\usepackage{threeparttable}
\usepackage{hhline}
\usepackage{makecell}
\usepackage{subcaption}
\usepackage{siunitx}
\usepackage{multirow}
\usepackage{boldline}
\usepackage{footmisc}
\usepackage{tabto}
\usepackage{amsfonts} 
\usepackage{balance}
\usepackage[belowskip=-1.0em,aboveskip=0em]{caption}
\usepackage{slashbox}
\usepackage[dvipsnames]{xcolor}
\usepackage{hyperref}
\usepackage{comment}
\usepackage{cite}
\usepackage{soul} 
\title{\LARGE \bf

}

\title{\LARGE \bf
Dynamic Bipedal MPC with Foot-level Obstacle Avoidance and Adjustable Step Timing

\author{ Tianze Wang and Christian Hubicki} 
\thanks{*Toyota Research Institute, Ford Motor Company, and the Mechanical Engineering Department at the FAMU-FSU College of Engineering provided funds to support this work.}
\thanks{$^{1}$All authors are with the Department of Mechanical Engineering,
         Florida State University, FAMU-FSU College of Engineering, Tallahassee, FL 32310, USA.
        Corresponding author: {\tt\small tw19j@fsu.edu}}
}

\begin{document}

\maketitle
 \thispagestyle{empty}
 \pagestyle{empty}

\begin{abstract} 
\\
Collision-free planning is essential for bipedal robots operating within unstructured environments. This paper presents a real-time Model Predictive Control (MPC) framework that addresses both body and foot avoidance for dynamic bipedal robots. Our contribution is two-fold: we introduce (1) a novel formulation for adjusting step timing to facilitate faster body avoidance and (2) a novel 3D foot-avoidance formulation that implicitly selects swing trajectories and footholds that either steps over or navigate around obstacles with awareness of Center of Mass (COM) dynamics. We achieve body avoidance by applying a half-space relaxation of the safe region but introduce a switching heuristic based on tracking error to detect a need to change foot-timing schedules. To enable foot avoidance and viable landing footholds on all sides of foot-level obstacles, we decompose the non-convex safe region on the ground into several convex polygons and use Mixed-Integer Quadratic Programming to determine the optimal candidate. We found that introducing a soft minimum-travel-distance constraint is effective in preventing the MPC from being trapped in local minima that can stall half-space relaxation methods behind obstacles. We demonstrated the proposed algorithms on multibody simulations on the bipedal robot platforms, Cassie and Digit, as well as hardware experiments on Digit.

\end{abstract}

\section{Introduction}

\begin{figure}[t]
    \centering
    \includegraphics[scale=1.0]{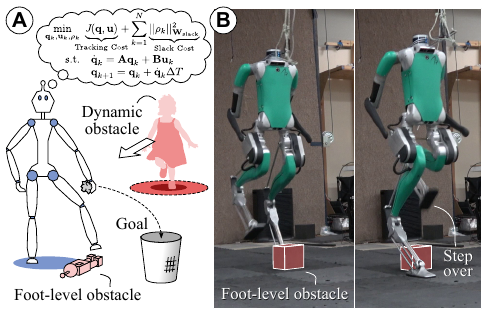}
    \caption{\textbf{(A)} Bipedal robots must avoid dynamic and foot-level obstacles in their environment. \textbf{(B)} A Digit humanoid using model-predictive control (MPC) to step over foot-level obstacles.}
    \label{fig:Intro}
\end{figure}

Navigation in human-populated environments is a challenging task for humanoid robots. To operate safely in real-world settings, bipedal robots must avoid collisions with various obstacles, including people, moving objects at different speeds, and obstacles on the ground (Fig.~\ref{fig:Intro}A). Therefore, it is necessary to a design control framework that reacts to obstacles with different locations and dynamics. 
Optimization-based control has demonstrated effectiveness in achieving locomotion and navigation tasks simultaneously by balancing different objectives~\cite{wensing2023optimization}. However, the optimal behaviors are also implicitly shaped by parameters such as target walking velocity or step frequency, which are typically hand-tuned. Introducing intelligent adjustments of these parameters can significantly reduce reliance on an engineer's expertise, thereby enhancing operation in complex environments. In this paper, we found that obstacle avoidance is significantly improved for bipeds if allowed to flexibly \textit{adjust step timing} to avoid fast-moving obstacles and selectively \textit{relax velocity tracking} when stepping over foot-level obstacles -- decisions which are novelly embedded in our convex model-based control formulation (Fig.~\ref{fig:Intro}B).

\subsection{Related Work}

Collision-free planning is an extensively researched area across a wide range of robotic platforms including mobile robots \cite{srinivasan2020control,9636406}, drones \cite{9981268,Marcucci2022MotionPA}, and legged robots \cite{gaertner2021collision,gilroy2021autonomous,inproceedings,10008227,shamsah2024socially,7940036}. For legged robots, safe navigation is particularly  challenging as it requires not only finding a collision-free trajectory but also stabilizing the robot around the planned path. This is due to the hybrid, nonlinear, and high degree-of-freedom dynamics inherent to legged locomotion.

Therefore, various approaches have addressed this challenge by integrating bipedal dynamics into motion planning. Specifically, reduced-order models (ROMs) based on inverted pendulum \cite{973365,9560821,SLIP} are widely used in real-time trajectory planning due to their simplicity and capability of capturing essential dynamics in legged locomotion. Various techniques have been presented to achieve collision-free planning in both static environments~\cite{Li2021VisionAidedAN,9447796} and dynamic environments~\cite{8206415,shamsah2024socially, 10341951, 9878052} with ROM-based dynamics. However, such methods primarily focus on base avoidance. Additionally, the dynamic environments contained mostly slowly-moving objects so avoidance strategies for high-speed obstacles have yet to be explored. One strategy to enhance locomotion agility is to dynamically adjust the step time. Multiple studies have demonstrated that adapting step time improves stability during push recovery. In \cite{khadiv2020walking}, the divergent component of motion (DCM) is used to change the step frequency  where \cite{li2024adapting} employs an NMPC to predict the unstable component. However, such methods rely on detecting instability or applying a strong regularization to the foot pattern, which does not suit obstacle avoidance tasks. Therefore, a different metric is required.

Motion planning with constrained footholds is another active research field. While distance based constraints can be included directly to enable foot collision avoidance\cite{hildebrandt2019kinematic,7743516,gaertner2021collision} along with methods like control barrier function (CBFs)~\cite{Agrawal2017DiscreteCB,khazoom2024tailoring}, the space is generally decomposed into several collision-free convex polygons to facilitate computation~\cite{9035046,9981419,acosta2023bipedal,7041373,song2021solving}. In~\cite{7041373,song2021solving}, Mixed-Integer Quadratic Programming (MIQP) is utilized to select footholds from a set of valid convex regions. However, those methods rely heavily on the quasistatic stability of the bipedal platform. Recently, \cite{acosta2023bipedal} achieved planning with constrained footholds on underactuated bipedal robots, where the 3D swing foot trajectory is generated by constructing splines rather than being optimized by MPC. 


In this work, we aim to design an MPC framework based on ROM that enables both agile body obstacle avoidance and collision-free planning for swing feet. Additionally, the framework can dynamically adjust commonly user-defined parameters, such as walking velocity and step frequency, based on feedback from the MPC solution, thereby reducing the reliance on human operators. Finally, we seek to demonstrate that this motion planning approach is practical for whole-body bipedal controllers, such as Operational Space Control~\cite{Apgar2018FastOT}.
\subsection{Contributions}
The contributions of this paper are:

\begin{itemize}
    \item An MPC framework for bipedal obstacle avoidance for both body and feet.
    \item A novel algorithm and heuristic for adjusting foot-step timing to increase avoidance agility for high-speed obstacles.
    \item A novel foot avoidance MPC formulation that implicitly chooses to step over or navigate around obstacles.
    \item Validation on bipedal robots on a multibody simulation of the Cassie biped and hardware avoidance experiments with the Digit biped.
\end{itemize}

To our knowledge, this is the first documented model-based approach for 3D dynamic bipedal walking that implicitly chooses to step over or navigate around foot-level obstacles depending on the state of its CoM dynamics.

\label{sec:introduction}

\section{Methods}
\begin{figure}[t]
    \centering
    \includegraphics[scale=1.0]{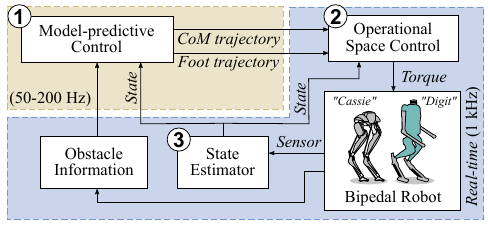}
    \caption{Block diagram for the presented controller. \textbf{(1)} MPC computes a collision-free CoM and Feet trajectory at 50-200Hz. \textbf{(2)} The Operational Space Controller computes the torques necessary to track the MPC reference at 1kHz. \textbf{(3)} An EKF is implemented to estimate base states.}
    \label{fig: Cassie Block Diagram}
\end{figure}

Here we present our method of generating dynamic bipedal motion plans that avoid body and foot collisions using MPC, shown in Fig. \ref{fig: Cassie Block Diagram}. We briefly introduce our previous formulation~\cite{10341951} in Sec. \ref{sec: recap} as the baseline controller. The method is extended in Sec. \ref{sec: adaptive step time} to accommodate avoidance of high-speed dynamic obstacles. Finally, we detail our foot-level obstacle avoidance formulation in Sec. \ref{sec: foot avoidance}.


\subsection{Conventions}
The following notation is used throughout the paper: $\mathbb{R}, \mathbb{R}^{n}, \mathbb{R}^{n\times m}$ represent the space of real numbers, vectors with length $n$, and matrices with $n$ rows and $m$ columns. Scalar values are lowercase and italicized letters (e.g. $\textit{x}$). Bold, lowercase letters represent vectors (e.g. $ \textbf{q} \in \mathbb{R}^{n}$) while bold, uppercase letters represent matrices  (e.g. $ \textbf{A} \in \mathbb{R}^{n\times m}$). \\
  
\subsection{Model Predictive Control}
Model Predictive Control (MPC) is a real-time control technique that utilizes optimization tools to find the optimal future trajectories for a system's states and inputs by minimizing performance costs while ensuring operational constraints are met. The MPC formulation presented in this work takes the general form:
\begin{equation*}\label{QP_Formulation}
\begin{aligned}
    \underset{\mathbf{q}_{k}, \mathbf{u}_{k}, \rho_{k}, \mathbf{b}_{k}}{\text{min}} \quad &   \underbrace{\textit{J}(\mathbf{q,u})}_\text{Tracking Cost} + \sum_{k=1}^{N}{\underbrace{||\rho_k||_{\textbf{W}_\text{slack}}^2}_\text{Slack Cost}} \\
    \text{s.t.} \quad & \dot{\textbf{q}}_{k} = \textbf{A}\textbf{q}_k + \textbf{B}\textbf{u}_k &\textrm{(Dynamics)}\\
    & \textbf{q}_{k+1} = \textbf{q}_k + \dot{\textbf{q}}_k \Delta T  &\textrm{(Model Prediction)}\\
    & h_{\textrm{slack}}(\textbf{q}_k,\rho_k,\textbf{u}_{k}) \leq 0  &\textrm{(Slack  Constraints)}\\
    & h_{\textrm{task}}(\textbf{q}_k,\textbf{u}_k) \leq 0 &\textrm{(Task Constraints)}\\
    & g_{int}(\textbf{q}_k,\textbf{u}_k,\textbf{b}_k, \rho_{k}) \leq 0 &\textrm{(Integer Constraints)}
\end{aligned}
\end{equation*}
where the robot states $\textbf{q}$, control input $\textbf{u}$, binary variables $\textbf{b}$, and all slack variables $\mathbf{\rho}$ are the decision variables in the optimization. This formulation only includes convex quadratic costs and linear constraints to allow for fast convergence and real-time application expect for the non-convexity introduced by the binary variables. 

\subsection{MPC for Bipedal Base Avoidance} \label{sec: recap}
Here we briefly introduce our MPC formulation that plans collision-free trajectory for the body and refer readers to \cite{10341951}  for a detailed description. We begin by defining
\begin{itemize}
    \item $N_\text{s}$ as the number of predicted robot steps,
    \item $T_\text{step}$ as the duration of each step,
    \item $\Delta T$ as the discretization time of the MPC,
    \item $N_\text{r} = \frac{T_\text{step}}{\Delta T}$ as the number of MPC nodes in each robot step,
    \item $N = N_\text{s}N_\text{r}$ as the number of MPC prediction nodes.
\end{itemize}




\subsubsection{Velocity tracking and foot placement cost}
\hfill

The MPC formulation builds upon our previous work \cite{10341951}, which uses the Linear Inverted Pendulum Model (LIPM). However, we no longer regularize all nodes to track a target velocity for high-speed obstacles avoidance or large disturbances, as it is difficult to design a reference trajectory for such tasks. To improve flexibility and avoid over-constraining the motion, we only penalize velocity tracking errors and deviations from the nominal foot-to-pelvis distance during step transitions. This promotes periodic stability overall without restricting motion within steps. Although better models are needed for optimal prediction and avoidance, our new formulation improves performance in experiments without added complexity.

Here, we detail the new formulation in the sagittal plane while a similar procedure can be applied to the frontal plane.

The analytical solution of LIPM in the sagittal plane is:
\begin{equation}
\begin{aligned}
    x(t) &= c_1 e^{\omega_0 t} + c_2 e^{-\omega_0 t} + \Bar{u}_{\text{x}}, \\
    \dot{x}(t) &= c_1\omega_0 e^{\omega_0 t} - c_2\omega_0 e^{-\omega_0 t}
\end{aligned}
\end{equation}
where $\omega_0 = \sqrt{\frac{g}{z_0}}$ is the natural frequency. Coefficients $c_1$, $c_2$ and foothold $\Bar{u}_x$ are free variables and we impose the following constraints to solve for them: (1) the trajectory needs to be periodic $\dot{x}(0) = \dot{x}(T_\text{step})$, (2) the trajectory should follow a desired velocity $\dot{x}(0) = \dot{x}_{\text{ref}}$, and (3) the initial condition $x(0) = 0$ given we only need the relative position between the foot and pelvis.
Once solved, we can define the tracking cost at step transitions (i.e. only MPC nodes at $i = kN_\text{r}$ for $k = 1, ..., N_\text{s}$) 
\begin{equation} 
\begin{aligned}
    \textit{J}_{\text{state}} &= \sum_{k=1}^{N_\text{s}} ||\dot{x}_{kN_\text{r}} - \dot{x}_{\text{ref}}||_{\textbf{W}_{\text{state}}}^2,\\
    \textit{J}_{\text{foot}} &= \sum_{k=1}^{N_\text{s}} || x_{kN_\text{r}} - u_{\text{foot}, kN_\text{r}} - \Bar{u}_\text{x}||_{\textbf{W}_{\text{foot}}}^2 .
\end{aligned}
\end{equation}

For nodes that are not at step transitions, the velocity tracking cost is zero and the footholds remain the same within each step (i.e., $u_{\text{foot}, \text{i}} = u_{\text{foot}, {kN_\text{r}}}$ for $(k-1)\text{N}_\text{r} < i \leq kN_\text{r}$). Finally, we introduce a control input tracking cost on footholds $\textbf{u}_\text{foot}$ which enables the MPC to better account for the control authority provided by the leg motors:

\begin{equation} 
    \textit{J}_{\text{effort}} = \sum_{k=1}^{N} ||\textbf{u}_{k} - \textbf{u}_{\text{foot},k}||_{\textbf{W}_{\text{effort}}}^2
\end{equation}

\subsubsection{Avoidance Constraint and Cost} \label{sec:base avoidance}
\hfill

\begin{figure}[t]
\centering
\includegraphics[scale=1]{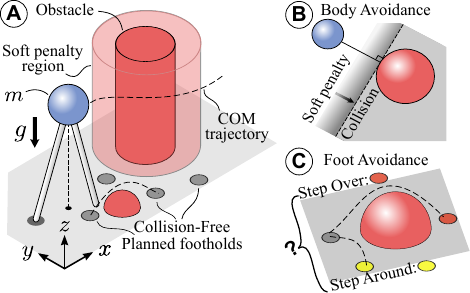}
\caption{\textbf{(A)} A LIPM-based MPC is implemented to generate collision-free trajectories. \textbf{(B)} The approximation is improved by linearizing constraints around a trajectory from the previous MPC solution. \textbf{(C)} We used a half-space relaxation with slack variables to model the collision-free regions~\cite{10341951}.}
\label{fig:LIPM Half-Space}
\end{figure}

We use the linear approximation method for body obstacle avoidance as in\cite{10341951} which divides the space into two regions and restricts the agent to operate within only one of them, illustrated in Fig.~\ref{fig:LIPM Half-Space}.

\subsection{Adaptive Stepping Time with Fast Moving Obstacles} \label{sec: adaptive step time}
In this subsection, we modified the MPC so it can avoid fast moving obstacles by refining the half-space approximation and introducing adaptive step frequency.

\subsubsection{Ellipse based Half-Space Formulation}

Here we introduce an alternative method for computing the normal vector $\hat{\textbf{d}}_{k}$. The previous approach based on linearizing the Euclidean distance falls short when dealing with high-speed obstacles because it relies solely on positional information. By incorporating the obstacle's velocity information, the MPC can select an avoidance path that is optimal not only for the obstacle's current position but also for its future trajectory.

To incorporate velocity information, we model the collision region as an ellipse, with the axis length determined by the obstacle's future path that is computed based on the obstacle's size, position, and current velocity. We then compute the point on the ellipse that is closest to the robot and use the tangent vector at this point as the avoidance direction. We found the closest point by solving the following optimization problem:
\begin{equation}\label{eq: Ellipse}
\begin{aligned}
    \underset{\mathbf{p}}{\text{min}} \quad &  ||\textbf{p} - \textbf{p}_{\text{0}}||^2 \\
    \text{s.t.} \quad & \frac{(\textbf{p}_{\text{x}}-\textbf{p}_{\text{obs,x}})^2}{a^2} + \frac{(\textbf{p}_{\text{y}}-\textbf{p}_{\text{obs,y}})^2}{b^2} = 1
\end{aligned}
\end{equation}
where $a, b$ are the axis lengths. Note, this new approach is same as the previous one when $a = b$. In Fig.~\ref{fig: step time}A, we use both methods to compute the avoidance direction with the same configuration and show that the direction computed by the new approach is more effective.

\subsubsection{Adaptive Robot Step Time}

The avoidance performance is also limited by the constant robot step time typically used in model-based bipedal controllers. Allowing the controller to dynamically adjust the step time could significantly enhance performance, especially  when the robot needs to execute lateral dodging maneuvers. This is because the lateral motion is notably restricted during the single stance phase, as the robot can only fall away from the stance foot which presents a challenge when an obstacle approaches from the side of the swing foot. Consequently, allowing the robot to adapt its step timing could better exploit the stance leg that facilitates dodging maneuvers.

\begin{figure}[tb]
\centering
\includegraphics[scale=1]{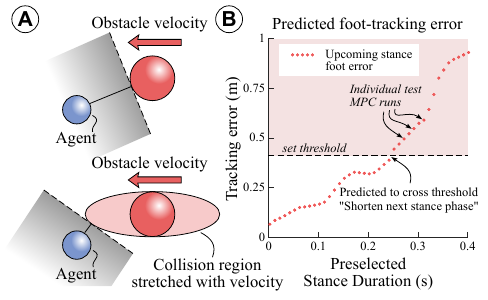}
\caption{\textbf{(A)} The avoidance direction computed based on the ellipse-based method is more effective by considering obstacle's velocity. \textbf{(B)} The step timing of the next step can be adjusted based on the MPC foothold tracking error.}
\label{fig: step time}
\end{figure}

Interestingly, we found that the error in center of pressure/foothold tracking can not only trigger a standing/stepping switch, as demonstrated in our previous work\cite{10341951}, but also serve as a criterion for adjusting the step time. To illustrate the step time switching criteria, we solve multiple MPC instances offline with step duration as the only changing parameter where the next stance phase would limit avoidance of an impending collision. We then plot the normalized tracking errors in foothold for each choice of stance time.  As shown in Fig.~\ref{fig: step time}B, the next stance foot exhibits higher errors as step timing increases. This observation aligns with intuition, as it is preferable to spend less time on the stance foot that constrains dodging. From an MPC perspective, this behavior arises because the avoidance cost accumulates significantly and outweighs the foot tracking costs when an inconvenient stance foot is selected.

Treating step time as a continuous variable is computationally intractable in real-time, so we limit the step time to several discrete choices at $0.1$s intervals. Accordingly, the step time of the next stance foot is reduced by $0.1$s if the tracking error exceeds a threshold until it reaches a minimum duration. Even with this quantized approach we notice a substantial improvement in the dodging behavior.

\subsection{Foot Avoidance} \label{sec: foot avoidance} 
In this section, we extend our base avoidance formulation to foot avoidance, enabling the robot to either navigate around or step over obstacles on the ground.

\subsubsection{Foot Avoidance with Mixed-Integer Quadratic Programming} \label{sec: MIQP}

The half-space formulation presented previously is not sufficient for foot avoidance because bipedal robots can step over the obstacles, whereas the half-space method simply divides the space in half and restricts movement to one side. If not selected correctly, the half-space could push the swing foot away from the intended path. To address this limitation, we utilize mixed-integer programming to enable planning with multiple admissible safe regions. As shown in \cite{acosta2023bipedal}, the MIQP can be solved relatively efficiently ($\sim$50Hz) with modern numerical solvers like Gurobi\cite{gurobi}.

\begin{figure}[tb]
\centering
\includegraphics[scale=1.0]{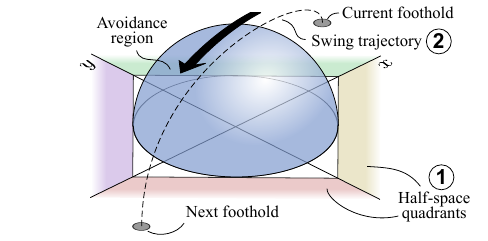}
\caption{Illustration of our swing foot 3D obstacle avoidance: \textbf{(1)} We use Mixed Integer Quadratic Programming to select foothold from the decomposed safe region represented by a quadrant. \textbf{(2)} Height constraints is imposed along the swing foot trajectory to achieve collision avoidance in 3D.}
\label{fig: Quadrant}
\end{figure}

To simplify the problem, we decompose the 3D avoidance into separate components for the $x,y$ plane and the $z$ plane, as shown in Fig.~\ref{fig: Quadrant}.  In the $x,y$ plane, the MPC only constrains the swing foot to start and end in a collision-free region. We then connect the endpoints with a straight line and impose height constraints for every nodes along this line to respect the obstacle's known position and size. Finally, we include slack-variable-based avoidance constraints and costs to each nodes in a manner similar to Sec.~\ref{sec:base avoidance} and detailed here for completeness. We first introduce the following variables: binary variables $\textit{b}_{k,i}$ for $i = 1,2,3,4$, position variables for the current swing foot $f_{\text{x},\text{i}}, f_{\text{y},\text{i}}, f_{\text{z},\text{i}}$ for $i = 1,...,N_r$, and foot avoidance slack variables $\rho_{\text{foot},\text{x}}$, $\rho_{\text{foot},\text{y}}$, $\rho_{\text{foot},\text{z}}$ into the MPC shown in Sec.~\ref{QP_Formulation}.

For the swing foot landing position in the $x,y$ plane, we assume the ground is divided into quadrants by the obstacle, as depicted in Fig.~\ref{fig: Quadrant}. Each region is defined as a set of linear constraints, $F_i u_k \leq c_i$, and is associated with a binary variable (e.g. $\textbf{b}_{k,1} = 1$ means foothold, $k$, is within the first candidate region). Similar to \cite{acosta2023bipedal}, we include the following constraints:

\begin{equation} \label{eq: region select}
\sum_{i=1}^{4} \textit{b}_{\text{k},\text{i}} = 1, \quad k = 1,...,N
\end{equation}

\begin{equation} \label{eq: region constraint}
F_\text{i} u_\text{k} + \rho_{\text{foot},\text{xy}, i} \leq c_\text{i} + (1 - b_{\text{i},\text{k}}) M, \quad i = 1,...,4
\end{equation}
where Eq.~(\ref{eq: region select}) implies exactly one candidate region must be active. Constraint (\ref{eq: region constraint}) is a technique called the big-M formulation. For a sufficiently large M, the i-th constraint (\ref{eq: region constraint}) is only active when $b_{\text{i},\text{k}} = 1$ and relaxed otherwise. 

Step height constraints are also enforced using the height profile $h_\text{i}$ derived from the geometry of the obstacle, per
\begin{equation} \label{eq: swing height}
    \textit{f}_{\text{z},\text{i}} + \rho_{\text{foot},\text{z}, i} \geq h_\text{i}.
\end{equation}
We then penalize all the slack variables $\rho_{\text{foot}}$ to avoid collision where we tune the weight $\text{W}_\text{foot,avoid}$ in each direction:
\begin{equation} \label{eq: swing slack}
    \textit{J}_{\text{foot,avoid}} = {\text{W}_\text{foot,avoid}}\rho_{\text{foot},\text{xyz}}^2.
\end{equation}

Finally, the swing foot should follow a reference trajectory $\textit{f}_{\text{ref},\text{k}}$ which is constantly updated based on the previous MPC solution:
\begin{equation} 
    \textit{J}_{\text{swing}} = \sum_{k=1}^{N_r} ||\textit{f}_{\text{k}} - \textit{f}_{\text{ref},\text{k}}||_{\textbf{W}_{\text{swing}}}^2
\end{equation}

\hfill

\subsubsection{MIQP Modification to Avoid Local Minima}

Although the MIQP formulation presented in Sec.~\ref{sec: MIQP} appears effective, we notice potential issue that may cause the MPC to become trapped in a local minimum that never traverses an obstacle, specifically due to the following factors:

\begin{itemize} 
\item There is a trade-off between minimizing the velocity tracking cost and the obstacle avoidance cost.
\item The prediction horizon is limited due to the computational demands of real-time control. This implies that passing the obstacle is optimal only when it is close enough. Otherwise, the MPC can only foresee the agent is approaching the obstacle which means slowing down is optimal due to the increases in the avoidance cost.
\item Moreover, the foot avoidance cost is symmetric and thus is the same whether the feet have overtaken or not overtaken the obstacle. This also leads to the biped stalling and never traversing the obstacle.  
\end{itemize}

These issues can be theoretically mitigated by making bespoke adjustments to the velocity command made by an experienced operator, implementing a higher-level planner with an extended prediction horizon or introducing asymmetry in the avoidance cost. However, we propose an extensible approach that involves the following additional costs and constraints to encourage the MPC to prioritize adjusting the walking speed and direction:
\begin{equation} \label{eq: dis cost}
\begin{aligned}
    \textit{J}_{\text{state,x}} &= \sum_{k=1}^{N_\text{s}} ||\dot{x}_{kN_\text{r}} - \dot{x}_{\text{ref},{kN_\text{r}}} - \rho_{\text{dis}}||_{\textbf{W}_{\text{state}}}^2,\\
    \textit{J}_{\text{state,y}} &= \sum_{k=1}^{N_\text{s}} ||\dot{y}_{kN_\text{r}} - \dot{y}_{\text{ref},{kN_\text{r}}} - \rho_{\text{dis}}||_{\textbf{W}_{\text{state}}}^2,\\
    \textit{J}_{\text{dis}} &= {\text{W}_{\text{dis}}}\rho_{\text{dis}} ^2
\end{aligned}
\end{equation}

\begin{equation} \label{eq: dis cons}
    \rho_{\text{dis}} + K_{\text{x}}x_{\text{dis}} + K_{\text{y}}y_{\text{dis}} \geq |\dot{x}_{\text{ref}}|T_{\text{step}} + |\dot{y}_{\text{ref}}|T_{\text{step}}
\end{equation}
where $x_{\text{dis}}, \, y_{\text{dis}}$ represent the predicted walking distances in the MPC and $K_{\text{x}}, \, K_{\text{y}}$ are tunable gains that prioritize movement in specific directions. Constraint (\ref{eq: dis cons}) enforces a minimum walking distance based on the reference velocity while still allowing flexibility in exploring different walking directions by balancing $x_{\text{dis}}$ and $y_{\text{dis}}$. Furthermore, the slack variable $\rho_{\text{dis}}$ becomes non-zero and stimulates a faster walking speed by changing the reference in cost (\ref{eq: dis cost}) when the travel distance requirements cannot be satisfied. By combining cost (\ref{eq: dis cost}) and constraint (\ref{eq: dis cons}), the MPC is forced to travel a minimum distance and encouraged to explore walking directions that might have lower long-term avoidance costs.

Note the non-linear distance term $x_{\text{dis}}$ and $y_{\text{dis}}$ can be formulated as linear constraints if we assume 1-norm distance and introduce additional binary variables.
\begin{equation} \label{eq: distance binary}
\begin{aligned}
    y &= y_{+} - y_{-} \\
    0 &\leq y_{+} \leq bM\\
    0 &\leq y_{-} \leq (1-b)M
\end{aligned}
\end{equation}
where $y$ is a real number represents the difference between the start and end position, $y_{+}$ and $y_{-}$ represent the positive and negative part of variable $y$, binary variable $b$ constrains positive and negative parts can not be both active. Then, the absolute $|y|$ term can be simply replaced by $y_{+} + y_{-}$, which is a linear constraint.

\subsection{Operational Space Controller}
We use an inverse-dynamics based operational space controller to track the MPC planned trajectory. For legged robots, the tasks are generally defined as CoM tracking and foot position tracking. In this work, we use the same quadratic programming based OSC implemented in \cite{10341951}.

\section{Simulation Results}

\subsection{Cassie Biped and Digit Humanoid}

Cassie and Digit are bipedal robots developed by Agility Robotics. The floating-base model of Cassie and Digit have 20 and 28 DOFs respectively where Digit's arms contain 8 DOFs total. The two robots share a similar mechanical leg design that contains passive joints implemented through a four-bar linkage with a mechanical spring. We use the simulation tools provided by Agility Robotics, namely, MATLAB Simulink for Cassie and MuJoCo~\cite{todorov2012mujoco} for Digit.

\subsection{Controller Implementation}
Our bipedal control framework consists of three components: a state estimator, a trajectory generator, and a tracking controller (shown in Fig.~\ref{fig: Cassie Block Diagram}). In this paper, we focus on the Digit controller implementation and briefly discuss the Cassie controller when necessary.

\subsubsection{State Estimation}

A contact-aided Kalman filter based on \cite{hartley2020contact} was developed in Matlab to estimate the pelvis translational and rotation states for Cassie. The Digit states are provided by the Digit low-level-api from Agility Robotics.

\subsubsection{QP Solver}

We used the Gurobi solver \cite{gurobi}  for optimization problems with binary variables and OSQP \cite{osqp} for other QPs. 

\subsubsection{MPC Setup}

Both foot and base avoidance MPC are implemented in Matlab and transferred to C++ using Casadi \cite{Andersson2019}. We used the following default MPC parameters through all experiments where the MPC planning horizon is set to four steps ($N_s = 4$) and the default step time is $0.4$ seconds. Currently, we set the MPC discretization time to be $0.02$ second based on the solving time of Gurobi MIQP. The safe distance $r$ is set based on the size of the obstacle.

\subsection{Simulation Results}
\subsubsection{Cassie Avoidance with Adaptive Step Time}
\hfill
\begin{figure}[tb]
\centering
\includegraphics[scale=1.0]{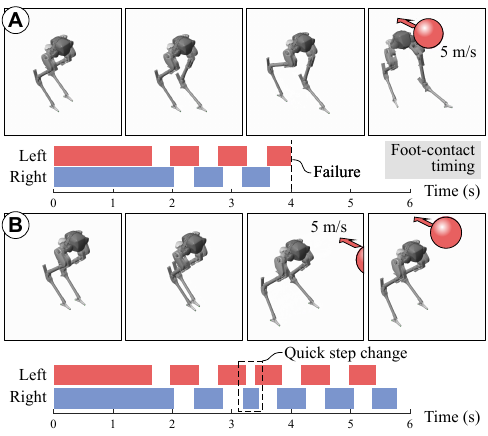}
\caption{\textbf{(A)} Avoidance failure from an impending 5 m/s obstacle is prevented by \textbf{(B)} utilizing obstacle's velocity information and introducing adjustable robot step time.}
\label{fig: Cassie_Sim}
\end{figure}

Here we present the Cassie avoidance simulation results, in which the robot attempts to avoid an approaching obstacle moving at 5 m/s. As a baseline, Fig.~\ref{fig: Cassie_Sim}A shows the result of the original formulation where Cassie fails the task due to the ineffective step timing. Fig.~\ref{fig: Cassie_Sim}B shows the result of including both the ellipse collision region formulation and adaptive robot step time in the MPC. During dodging maneuvers, the right step phase of Cassie is shortened to 0.2 seconds as necessary, allowing Cassie to transition to the left step phase more quickly which leads to a more effective dodging. MPC parameters such as obstacle's velocity and robot position remain the same through all experiments.

\subsubsection{Digit Foot Avoidance}
In the experiment, Digit is commanded to walk toward an obstacle with different command velocities. With a low commanded velocity, Digit navigates around the obstacle to avoid being trapped in local minima, while a higher commanded velocity saved time by stepping over the obstacle as shown in Fig.~\ref{fig: Digit_Hardware}A.
\section{Hardware Experiments} 
\subsection{Experimental Setup}
\subsubsection{Software Setup}

The Digit controller is coded on an Intel NUC which is connected to the Digit PC via Ethernet cable. Currently, the foot avoidance controller is implemented on a lab desktop due to computational demand from the MIQP. The controller is implemented in C++ with Robot Operating System (ROS) library\cite{ros}. The MPC is updated at 50Hz for MIQP and 200Hz otherwise.

\subsubsection{Hardware Controller Tuning}

We used the same controller in both simulation and hardware experiments and made the following modifications for Digit hardware control:
\begin{itemize}
    \item The Digit control interface requires three input commands: feed-forward torques, reference motor velocity and reference motor damping. We integrate the OSC solution to get the reference velocity and tune the motor damping manually on hardware.
    \item A small offset term on the order of $0.01$ is added to the foot position in the MPC to reduce the drifting issue caused by the model mismatch.
\end{itemize}

\subsubsection{Physical Setup}

The hardware experiments were performed on a Digit V3 humanoid from Agility Robotics in a laboratory environment with a catch rope only for safety purposes. The foot obstacle was positioned at a fixed location due to the absence of obstacle position sensing.

\subsubsection{Hardware Result: Digit Foot Avoidance}

\begin{figure}[tb]
\centering
\includegraphics[scale=1.0]{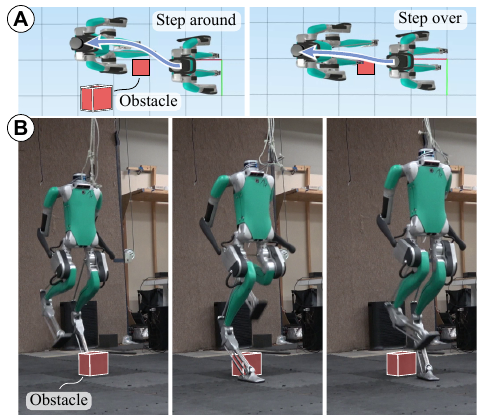}
\caption{\textbf{(A)} A simulation of Digit stepping around and over an obstacle respectively depending on its commanded speed. \textbf{(B)} Digit achieves 3D swing foot avoidance by stepping over the obstacle on hardware.}
\label{fig: Digit_Hardware}
\end{figure}

Here we show the Digit hardware experiments on foot avoidance. The size of the obstacle used in the experiments is $0.2$m and we set the diameter of the collision regions to be $0.4$m to account for the Digit foot length on both side. As shown in Fig.~\ref{fig: Digit_Hardware}B, Digit managed to adapt the walking velocity and step over the obstacles.

\section{Conclusion and Future Work}
In this paper, we present a novel motion-planning framework for bipedal robots, focusing on both bipedal body and foot avoidance. We extend our previous base avoidance formulation by incorporating an ellipse representation of the collision region which includes obstacle's velocity information and introducing an intelligent step time scheduling to improve the avoidance performance for high-speed obstacles. Foot avoidance is achieved by using Mixed-Integer Quadratic Programming, coupled with a slack variable-based minimum travel distance constraint to avoid being trapped in local minima. We validate the proposed framework through simulations on the Cassie bipedal robot, demonstrating adaptive step timing, and further evaluate its effectiveness with both base and foot avoidance on the Digit hardware platform. 

Future work will interface cameras on Digit to achieve onboard obstacle sensing, extending the framework to multiple ground obstacles with a more advanced decomposition method, introducing swing foot kinematics/dynamics via nonlinear MPC, and improving the robustness of the controller and conducting more experiments on hardware.
\section{Acknowledgment}
This work was supported by the Toyota Research Institute, the Ford Motor Company, and the Department of Mechanical Engineering at the FAMU-FSU College of Engineering. The authors acknowledge Jacob Hackett's assistance in reviewing the manuscript.
\balance
\bibliographystyle{IEEEtran}
\bibliography{literature}

\end{document}